\documentclass{article}

\usepackage{PRIMEarxiv}

\usepackage[utf8]{inputenc} 
\usepackage[T1]{fontenc}    
\usepackage{hyperref}       
\usepackage{url}            
\usepackage{booktabs}       
\usepackage{amsfonts}       
\usepackage{nicefrac}       
\usepackage{microtype}      
\usepackage{lipsum}
\usepackage{fancyhdr}       
\usepackage{graphicx}       
\graphicspath{{media/}}     
\usepackage[round]{natbib}

\usepackage{caption}
\usepackage{subcaption}
\usepackage{placeins}
\usepackage{hyperref}
\usepackage{amsmath}
\usepackage{amsthm}
\title{Uncertainty Quantification for Rule-Based Models
}

\author{
  Yusik Kim \\
  IBM Research \\
  \texttt{yusik.kim@ibm.com} \\
}

\begin{document}
\maketitle

\begin{abstract}
Rule-based classification models described in the language of logic directly predict boolean values, rather than modeling a probability and translating it into a prediction as done in statistical models. The vast majority of existing uncertainty quantification approaches rely on models providing continuous output not available to rule-based models. In this work, we propose an uncertainty quantification framework in the form of a meta-model that takes any binary classifier with binary output as a black box and estimates the prediction accuracy of that base model at a given input along with a level of confidence on that estimation. The confidence is based on how well that input region is explored and is designed to work in any OOD scenario. We demonstrate the usefulness of this uncertainty model by building an abstaining classifier powered by it and observing its performance in various scenarios.
\end{abstract}


\section{Introduction}
Rule-based classification models are constructed from rules having the structure ``if \textlangle condition\textrangle \;\;then \textlangle conclusion\textrangle''. 
These are widely used in industry for applications such as business process management, fraud detection, loan approval where model transparency, auditability, and the ability to directly encode domain knowledge and user intent is required.
In industries where rules are traditionally human-authored, there is a recent trend to induce such rules from data as part of business automation initiatives.
Replacing traditionally human-authored rules or manual processes with automatically induced rule-based models requires trust that the trained model is going to make ``common sense'' decisions.
This includes not being overly confident on predictions for outlier inputs, and acknowledging borderline cases that need to be escalated to a human expert.
One way a model could have this kind of situational- and self-awareness is by having a good associated uncertainty model, either natively or provided externally. 

Classifiers based on statistical models typically model the probability of observing the output given the input.
Well known models such as neural networks and logistic regression all model this conditional probability.
This probability acts as a metric of confidence and is finally translated into a prediction.
Note that it is possible for a classifier to achieve high prediction accuracy without necessarily modeling the probabilities accurately.
Such overconfidence is a well known problem for neural network classifiers \citep{pmlr-v70-guo17a} and statistical classification models for prediction in general \citep{VanCalster2019}.
An important issue for quantifying uncertainty for statistical classification models is, therefore, how well the modeled probabilities are calibrated to agree with the empirical frequencies.
Provided that the modeled probabilities are well-calibrated, it could effectively serve as a metric of uncertainty.

Rule-based classification models, on the other hand, are not stochastic models describing the conditional probability of the output given the input, but are rather deterministic models that directly model the prediction as the conclusion of a logical implication, i.e., an if-then rule.
The condition part of the rule involves a vocabulary of predicates (e.g., $x_i \geq 3$) combined with logical connectives (e.g., $\land, \lor, \neg$) to form logical expressions evaluating to a truth value.
Logical expressions being inherently binary, there is no obvious element of a rule-based model that we can use as a proxy for confidence as was possible for statistical models.

As calibrating the binary output of a rule-based model does not make sense, we consider an external uncertainty model, or a meta-model that models the prediction error rate of a given base classification model.
Note that this base model need not be rule-based, and our approach is equally applicable to any kind of classification model.
Given a base model, we model the conditional distribution of the error rate of its prediction at a given input point.
The goal is to model the distribution in such a way that 1) the variance is inverse-proportional to the density of input observations, and 2) the mean is well calibrated in the sense that it is close to the empirical error rate of the base model at each point of the input space.
This error model will allow us to make statements about how accurate we think the base model's prediction is (through the mean) and how certain we are about that statement (through the variance).
We provide in the sequel an interpretation of aleatory and epistemic uncertainties in this context, and how they can be represented through the mean and variance of the error model.

An immediate advantage of having a good uncertainty model is that we can make a principled decision on whether to abstain from accepting a prediction made by the base model.
We demonstrate what kind of improvement we can expect by abstaining in low-confidence situations in various out-of-distribution (OOD) scenarios. 

The scope of uncertainty discussed in this paper is limited to that of individual predictions made by a fixed classification model.
We acknowledge that there are other aspects of uncertainty that arise in different contexts.
For example, the kind of uncertainty involved in reasoning with probabilistic facts in probabilistic logic programming, or the uncertainty associated with estimated global performance of a model rather than at individual points under a different input distribution are beyond the scope of this paper.

The main contribution of this paper is a proposal of an uncertainty quantification framework that:
\begin{enumerate}
    \item quantifies the prediction uncertainty applicable to any binary classification model including rule-based models
    \item depends only on the binary output interface of the base model and not on any of its internal details
    \item is capable of providing uncertainty metrics for arbitrary points in the input space and is designed to work in OOD scenarios without any assumption on the input distribution.
\end{enumerate}

\section{Related Work}
Meta-models that model the uncertainty of an underlying base-model are not new.
There is a class of meta-models that work only with neural networks by accessing intermediate flow values from which a performance model of a base network model is built \citep{https://doi.org/10.48550/arxiv.1705.08500, pmlr-v89-chen19c, https://doi.org/10.48550/arxiv.1909.04079}. 
In addition to being limited to neural networks, these approaches are motivated as an alternative to calibration, and as such, there is no attempt to decompose the uncertainty into aleatory and epistemic components, and does not extend to OOD input.
Other meta-models that claim to be model agnostic still have the implicit assumption that the underlying base model is a statistical model, and attempts to provide interval estimates for the predictive distribution for a given input \citep{https://doi.org/10.48550/arxiv.1806.00550}.
\citet{DBLPabs-2106-00858} proposes a method for evaluating the quality of such prediction intervals and comparing them across models in a generic way.

For quantifying the uncertainty of the performance of an underlying base-model, \citet{https://doi.org/10.48550/arxiv.2012.08625} proposes a \emph{meta}-meta-model where a base prediction model and a meta-model that models the performance of the base model is given.
Then the meta-meta-model is used to get prediction intervals of the performance estimate provided by the meta-model.
Quantifying the uncertainty in unseen regions of the input space is done through simulated feature drift.
But crucially, it differs from our work in that it provides prediction intervals for the performance for entire datasets, not at individual data points.

A comprehensive framework on quantifying uncertainty for statistical models has been provided by \citet{DBLPabs-2011-07586}.
Although not directly applicable to rule-based models, the ideas presented there could potentially be applied to meta-models on the performance of a base model as we do in our work.
However, our interest lies in quantifying the uncertainty of the base model predictions with the help of the meta-model, not that of the meta-model itself.
A large body of work on conformal predictions \citep{DBLPabs-2107-07511} for predicting sets that contain the true class with high probability instead of point predictions for quantifying uncertainty is available, but they apply exclusively to statistical models providing continuous output.

In the classical rule learning literature, the confidence of a rule is equated with precision or its variants such as Laplace and $m$-estimates \citep{10.5555/2788240}.
As an entire data set is required to evaluate the precision, it is a static property of a rule, rather than a confidence metric that can be attributed separately to individual data points.
Also, the precision measured is only valid under the assumption that the input distribution does not change, so it does not extend to OOD scenarios.
Confirmation rule sets \citep{confrimationruleset} is a method that provides confidence at the data point level.
It represents prediction confidence by the number of rules that fired on a particular input.
However, it is a heuristic with no statistical or physical interpretation as a metric.

One of the advantages of building an abstaining classifier based on an uncertainty model that gives valid results over the entire input space is that it can naturally handle OOD scenarios.
There are many abstaining classifiers, for example \citet{Pietraszek05optimizingabstaining, DBLP:journals/corr/Balsubramani15a}, that do not support OOD scenarios, or \citet{DBLP:journals/corr/abs-2101-12523} that are applicable only to statistical models, or \citet{DBLP:journals/corr/abs-1901-09192} that works only for neural networks.
\citet{DBLP:journals/corr/abs-2105-14119} and \citet{DBLP:journals/corr/abs-2007-05145} propose abstaining classifiers that do support OOD scenarios but is limited to transductive learning settings where the solution is specific to a given training data and OOD test data pair, and does not generalize to arbitrary test data. These are also noteworthy as they don't even rely on defining a confidence metric or an uncertainty model but directly finds a set of examples to abstain on.

\section{Sources of uncertainty surrounding rule-based models}
A rule-based model of a possibly noisy binary output can be represented as a boolean regression model
\citep{Boros1995} with binary (bit) error:
\begin{align}
    y&=f(x)+\epsilon \mod{2}\nonumber\\
    \epsilon&\sim \mathrm{Bernoulli}(\pi)
    \label{eq:boolreg}
\end{align}
where $f$ is a logical formula and $\pi$ can be interpreted as the unknown bit error rate, which may be further modeled as a constant, a deterministic function of the input, or a random variable that depends on the input.

As a regression problem, the most general way to represent the relationship between the input $x$, the output $y$, and the (possibly noisy) environment $e$ is
\begin{align}
    y = \phi(x, e)
    \label{eq:ideal}
\end{align}
where we assume the existence of some true but unknown $\phi$ and $e$.
Provided that the specification of $\phi$ and $e$ completely determines the ideal model, the lack of knowledge about these two objects embodies any and all uncertainties concerning our problem.
In particular, the lack of knowledge about the true structure $\phi$ is what is commonly referred to as \emph{epistemic} uncertainty, and the irreducible randomness of the environment $e$ is what is referred to as \emph{aleatory} uncertainty.

For an instantiated model such as (\ref{eq:boolreg}) where $f$ is a deterministic logical expression with all randomness pushed outside as a bit error, the epistemic uncertainty amounts to how far away, structurally, $f(x)+\epsilon$ is from $\phi(x, e)$, and the aleatory uncertainty remains to be $e$.
Since both types of uncertainty eventually manifest themselves through the bit error $\epsilon$, we can ask how much of the randomness in $\epsilon$ can be attributed to the epistemic vs. aleatory uncertainties.
Although this decomposition is interesting from a model selection perspective, it is not useful for the purpose of quantifying the uncertainty on whether a prediction made by a given model is correct.
Once the base model is fixed, the randomness embodied by $\epsilon$ becomes effectively aleatory, as there is no way to further reduce it.
If we knew the true error rate $\pi$ at each point of the input space, there is no more knowledge we could hope to gain as far as prediction is concerned.
Therefore, the only remaining source of uncertainty is the lack of knowledge about the true error rate at a given point.

Not as a selector of the base model but as an observer of the predictions made by the base model, the aleatory uncertainty is represented by the Bernoulli error model with the true error rate, and the epistemic uncertainty is represented by the imperfect knowledge regarding the true error rate.
The missing piece for quantifying prediction uncertainties in this context is to find an appropriate model for $\pi$.
Through the model of $\pi$, we want to make an assessment on how accurate we expect the base model to be at a given input, and at the same time provide a level of confidence on that assessment.

When interpreting $\pi$ as a probability, rule-based models being deterministic can cause some problems.
For example, when the data is noise-free, meaning the same $x$ always has the same $y$, the error rate $\pi$ at a fixed $x$ must be either 0 or 1, as repeating the experiment would either always be right or always be wrong depending on whether the model happens to give the right prediction at that point. 
If we model $\pi$ in a way that it can only be estimated as a value strictly between 0 and 1, it is unclear how to interpret this as a probability in this case.
This is especially problematic since we said that the Bernoulli model with this estimated error rate will represent the aleatory uncertainty - something we do not have in a noise-free environment with a deterministic model.
The error rate model for $\pi$ we present in the sequel is far from perfect, and exhibits this problem. 
But just as logistic regression models on noise-free data can still be useful, we argue that our model is useful when interpreting the estimated error rates as a belief or a propensity to be wrong, rather than a frequency estimate of a repeated experiment.

The approach we take can be summarized as follows.
We consider a meta model that models the error made by a given base model.
The estimated error rate quantifies the aleatory uncertainty of the correctness of the prediction made by the base model.
The uncertainty on this estimated error rate represents the epistemic uncertainty.
The input to our approach are the (pre-trained) binary base classification model and any labeled data set. 
The output is a function that maps points of the input domain to individual distributions of the error rate at those points, i.e., a random field.

\section{Modeling the prediction error}
For a given base model $f$ that outputs boolean values (that are binary encoded), we model the observation $y$ in terms of the input $x$ using the following error model:
\begin{align}
\label{additivemodel}
    y&=f(x)+\epsilon \mod{2}\nonumber\\
    \epsilon&\sim \mathrm{Bernoulli}(\pi)\nonumber\\
    \pi|x&\sim ?
\end{align}
The task is to identify an appropriate model for $\pi|x$ that can quantify the aleatory and epistemic uncertainties associated with a prediction made by the base model $f$.
As discussed in the previous section, we view our estimate of $\pi|x$ as the (estimated) aleatory uncertainty, and our confidence on that estimate as the epistemic uncertainty.

Note that the data for training this prediction error model has no relation to the data used to train the base model, and consequently does not have to come from the same distribution.

\subsection{Modeling $\pi$ as a deterministic function}
Logistic regression is one of the simplest models we can consider for $\pi|x$.
The error rate can be estimated through the MLE $\hat{\beta}$ as 
\begin{align}
     \hat{\pi}|x=\mathrm{logistic}(\hat{\beta}^{\top}x).
\end{align}
Since the sampling distribution of $\hat{\beta}$ is asymptotically normal (property of the MLE), $\hat{\beta}^{\top}x$ is also asymptotically normal.
It follows that the standard error of $\hat{\beta}^{\top}x$ can be estimated and a confidence interval for $\beta^{\top}x$ can be constructed.
As the logistic function is monotonic and hence an order-preserving one-to-one map, the corresponding confidence interval for $\pi|x$ can be constructed.
The $\hat{\pi}|x$ is our estimate of the aleatory uncertainty of the base model, and the associated confidence interval represents the epistemic uncertainty of the model's prediction at a given input $x$.

A limitation with this approach is the model assumption that the propensity of the base model making an error monotonically increases in a certain direction of the input space. 
This issue can be addressed to some degree by techniques such as generalized linear models or polynomial regression.
But it is cumbersome to select an appropriate family of models, find the MLE, and estimate its standard error, perhaps through bootstrapping.
Another problem is regarding model misspecification.
The estimated standard error is only accurate when the assumptions for the logistic regression model hold in the data, namely that the error $\epsilon$ are independent given $x$ and they truly follow Bernoulli($\hat{\pi}|x$).
This is especially problematic when the observations are noise-free, as discussed in the previous section.
The true $\pi|x$ is either 0 or 1, but no confidence interval of it derived from $\hat{\beta}^{\top}x$ will ever be able to cover it.
A more serious problem with adopting a logistic regression model is that it does not have the desired property of reporting high epistemic uncertainty at points in the input space where observations are sparse. 
As long as the fit is good in the regions adequately observed during training, it unjustly extrapolates this confidence to regions with sparse or no observations.

\subsection{Modeling $\pi$ as a random function}
The simplest way to extend the logistic regression model to a random function is to endow a prior distribution over the $\beta$ parameters, i.e., a Bayesian logistic regression model:
\begin{align}
     \pi|x&=\sigma(\beta^{\top}x)\nonumber\\
     \beta&\sim \mathcal{N}(0, \Sigma)
\end{align}
where $\sigma$ is the logistic function.
This, however, does not address any of the three problems of the deterministic model.
Moreover, there is an additional problem introduced by employing a Bayesian approach: the posterior distribution of $\pi$ is a subjective belief based on the chosen prior.
While, in principle, the likelihood dominates the prior where there are many observations, the resulting posterior distribution of $\pi$ is generally not an accurate representation of what one could expect through empirical experiments.

Keeping this Bayesian approach but replacing the prior distribution over hyperplanes with a Gaussian process prior coupled with logistic likelihood \citep{10.5555/1162254} yields the model:
\begin{align}
\label{eq:GPerror}
     \pi|x&=\sigma(g(x))\nonumber\\
     g&\sim \mathcal{GP}(0, K).
\end{align}
So we are modeling the logit of the error rate
\begin{align}
    \log\frac{\pi}{1-\pi}
\end{align}
as a Gaussian process.
As a sampled $g$ is no longer structurally monotonic, we address the problem of monotonically increasing error rate of the deterministic model.
The issue of wanting the variance of the estimate to be large in areas with sparse observation can be achieved by using a high variance Gaussian process prior.
The issue of model misspecification is irrelevant in this case as the estimator is no longer the MLE.
However, as a Bayesian model, it suffers from the same problem regarding subjectivity as the Bayesian logistic regression model.
Although this is not ideal, if the purpose of using this quantified uncertainty is to simply make a decision on whether to accept a prediction made by a machine, this kind of relative/subjective assessments can still be useful if the acceptance thresholds need to be calibrated anyway.

\begin{figure*}[t]
     \centering
        \begin{subfigure}[t]{0.3\textwidth}
         \centering
         \includegraphics[width=\textwidth]{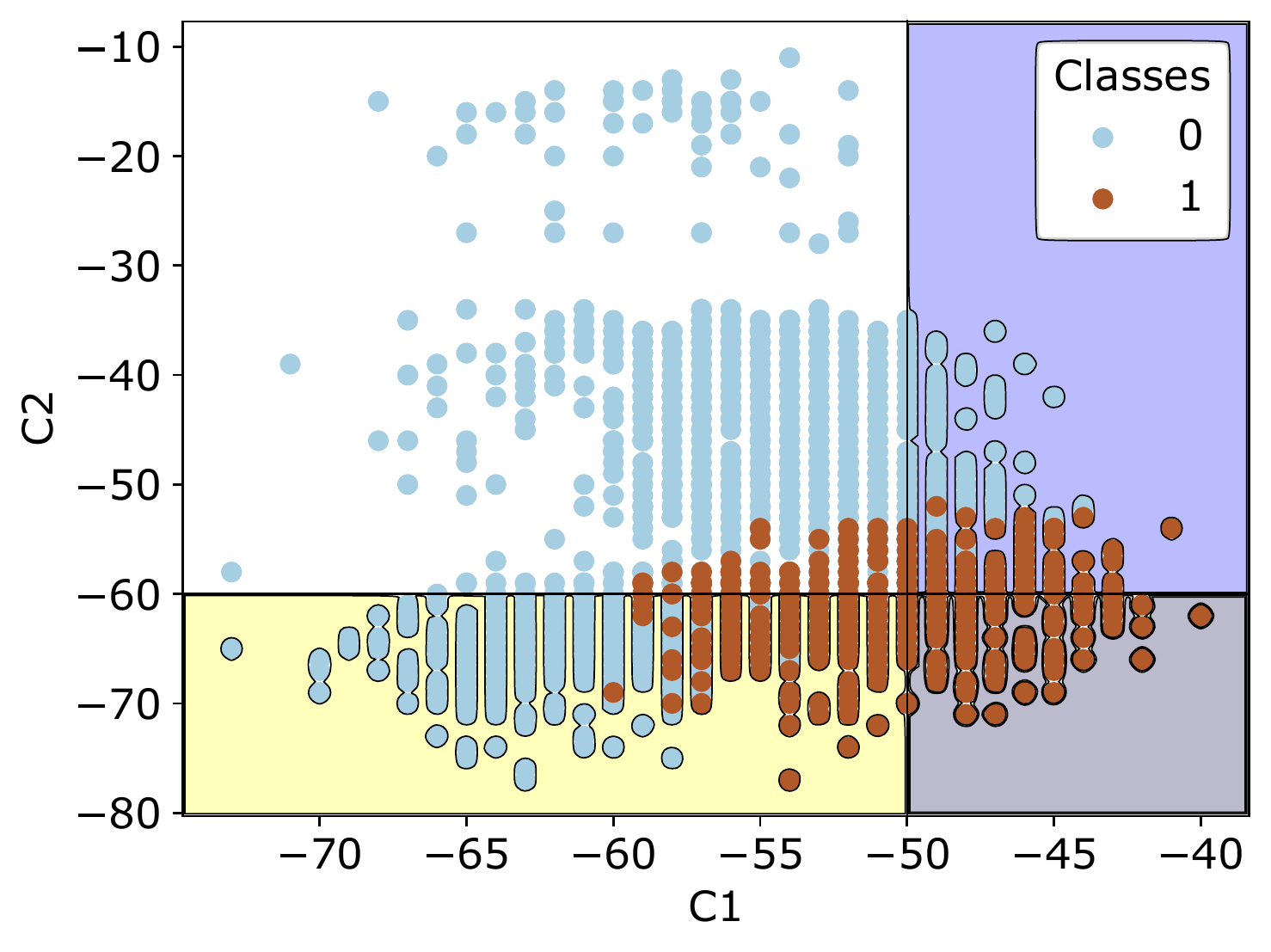}
         \caption{Scatter plot of the full data. The blue region on the right ($C1 > -50$) is where the training data for RIPPER is sampled. The yellow region on the bottom ($C2 < -60$) is where the training data for the GP error model is sampled.}
         \label{fig:scatter}
     \end{subfigure}
     \hfill
     \begin{subfigure}[t]{0.3\textwidth}
         \centering
         \includegraphics[width=\textwidth]{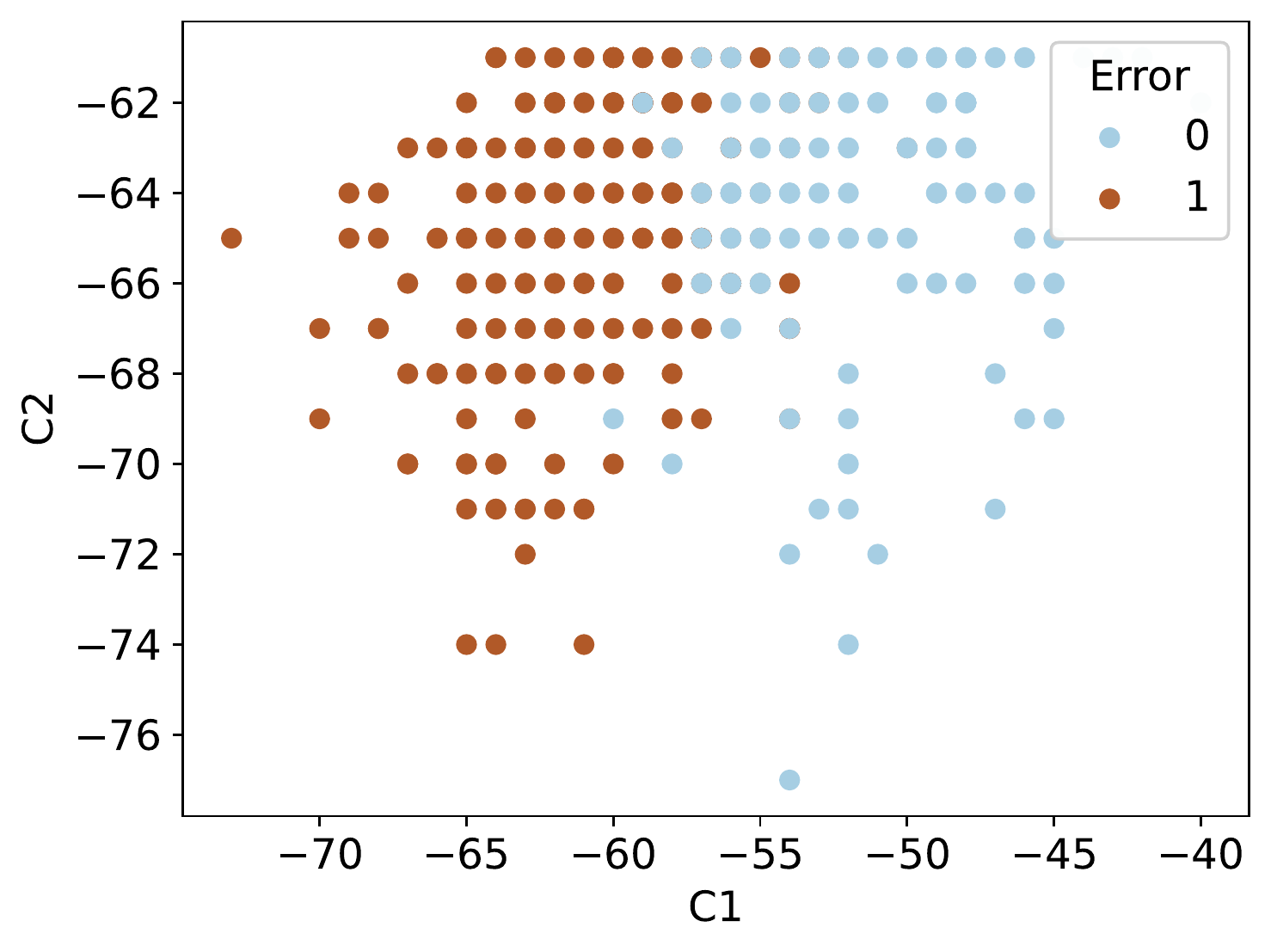}
         \caption{Scatter plot of the errors made by RIPPER in the yellow region.}
         \label{fig:error}
     \end{subfigure}
     \hfill
     \begin{subfigure}[t]{0.35\textwidth}
         \centering
         \includegraphics[width=\textwidth]{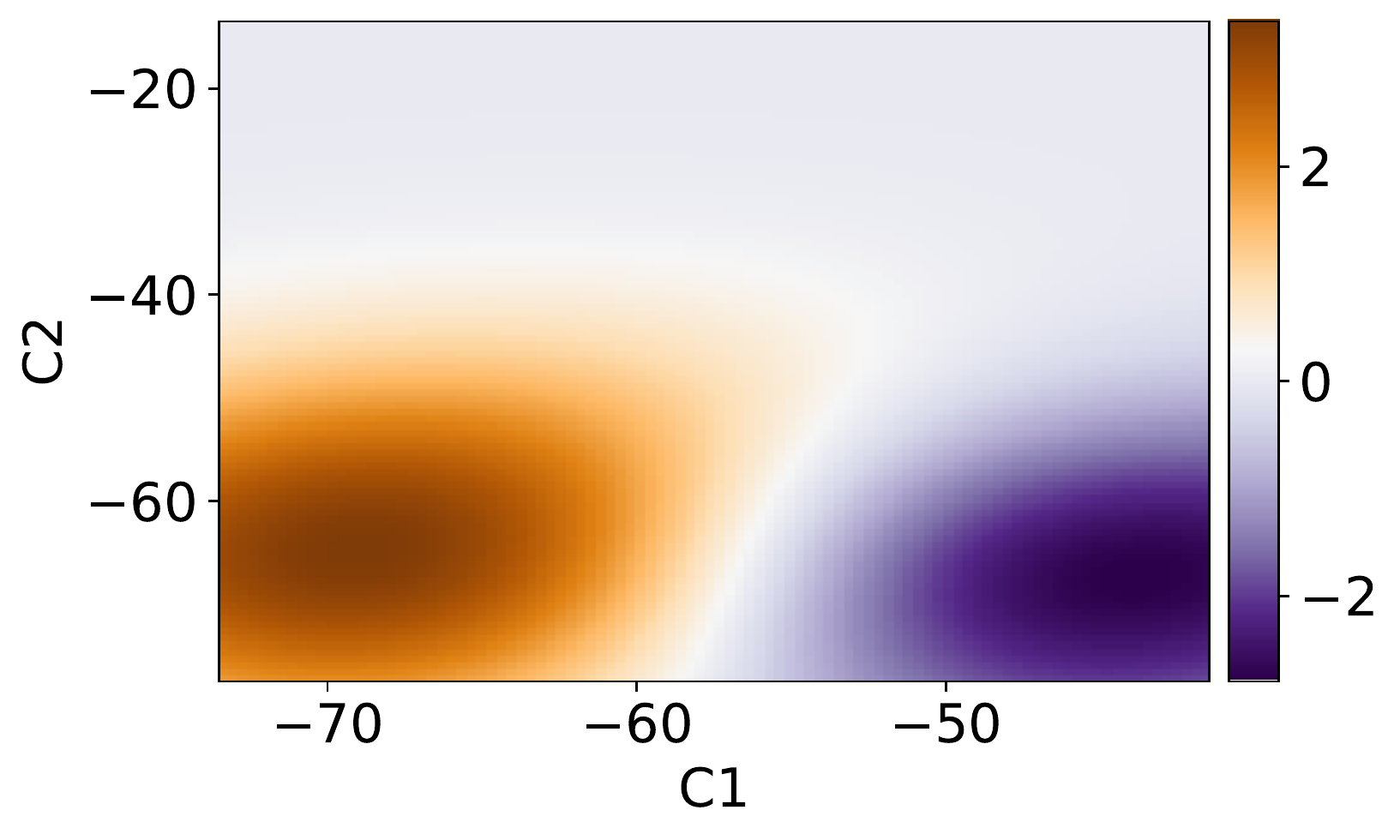}
         \caption{Posterior mean of the Gaussian process modeling the logit error rate given the observations of (a). Likely wrong prediction in red, likely right in blue, and unsure at the boundary and unobserved top region.}
         \label{fig:posterior_mean}
     \end{subfigure}
     \hfill
     \begin{subfigure}[t]{0.35\textwidth}
         \centering
         \includegraphics[width=\textwidth]{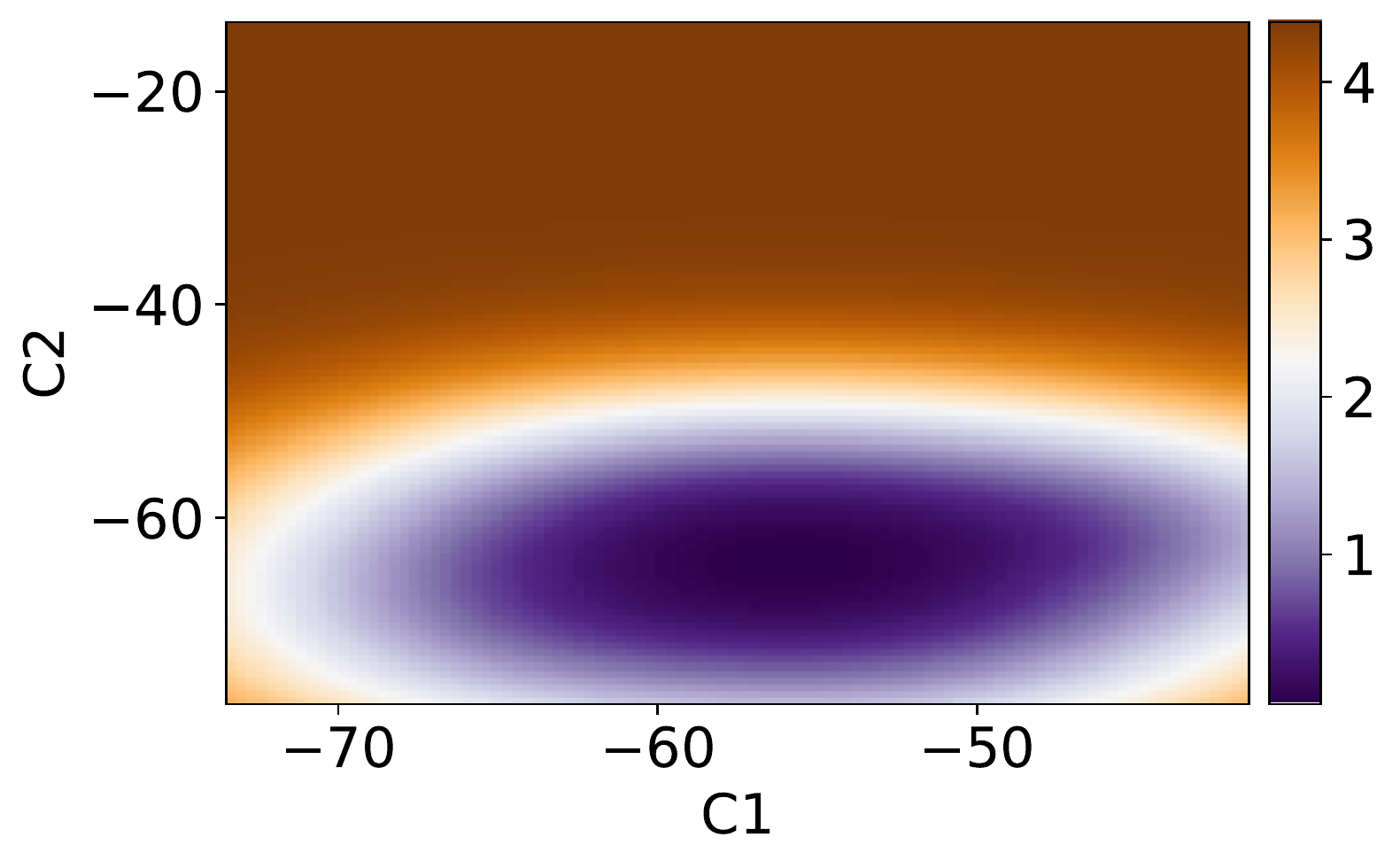}
         \caption{Posterior variance of the Gaussian process modeling the logit error rate given the observations of (a).}
         \label{fig:posterior_var}
     \end{subfigure}
     \hfill
          \begin{subfigure}[t]{0.3\textwidth}
         \centering
         \includegraphics[width=\textwidth]{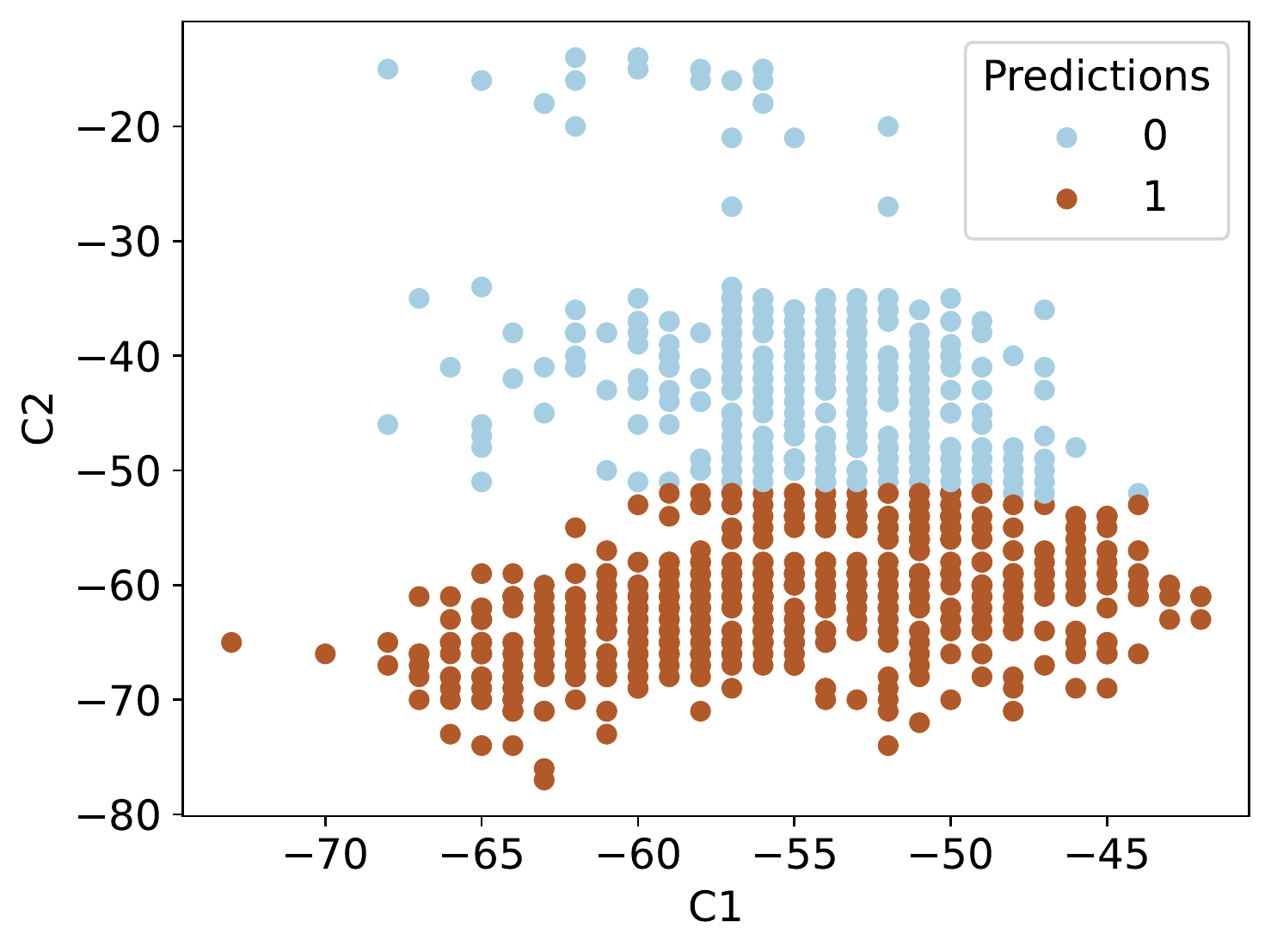}
         \caption{The predictions made by the RIPPER model on $S3$ without the help of the error model.}
         \label{fig:error2}
     \end{subfigure}
     \hfill
     \begin{subfigure}[t]{0.3\textwidth}
         \centering
         \includegraphics[width=\textwidth]{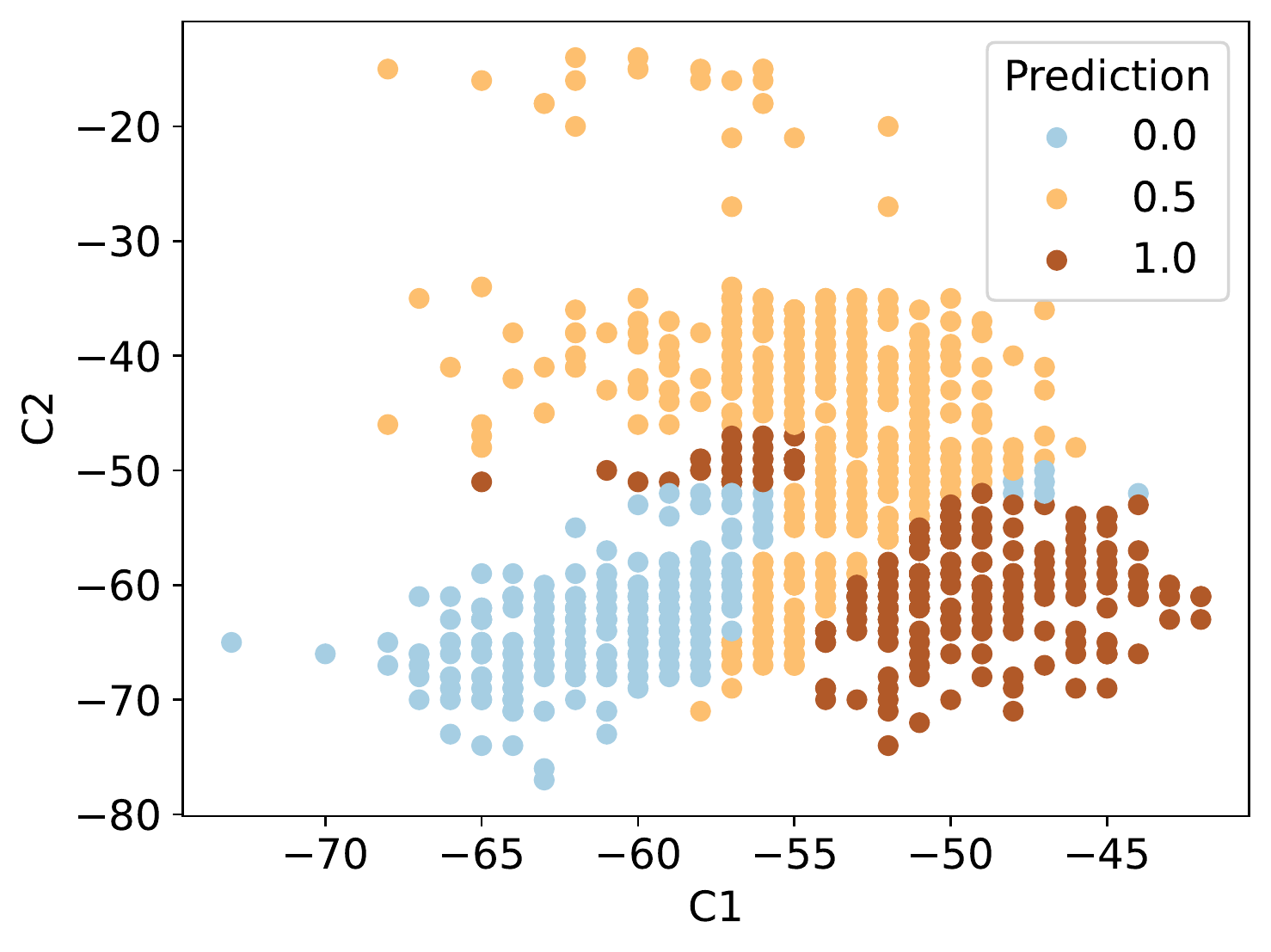}
         \caption{The predictions made by the RIPPER model on $S3$ with the help of the error model where the yellow points indicate abstention.}
         \label{fig:abstain}
     \end{subfigure}
        \caption{Connecting the data, the GP-based error model, and the resulting abstention decision.}
        \label{fig:three graphs}
\end{figure*}
Denoting the set of error observations of a given base model $f$ by $\mathcal{D}=\{(\epsilon_i, x_i) | i=1,...,N\}$, the posterior distribution of $g$ is
\begin{align}
    p(g|\mathcal{D})&\propto p(\mathcal{D}|g)p(g)\\
    &=p(g)\prod_i\sigma(g(x_i))^{\epsilon_i}(1-\sigma(g(x_i)))^{1-\epsilon_i}.
\end{align}
The predicted error rate at a new input $x_*$ is distributed as $\sigma(g(x_*))|\mathcal{D}$.
The posterior mean and variance of this Gaussian process can respectively be used as our metric for aleatory and epistemic uncertainty of the prediction at $x_*$:
\begin{align*}
    E[g(x_*)|\mathcal{D}]&&(\text{Estimate of the logit error rate at }x_*)\\
    \mathrm{Var}(g(x_*)|\mathcal{D})&&(\text{Confidence on the estimate at }x_*)
\end{align*}
As the likelihood $p(\mathcal{D}|g)$ is non-normal, the posterior does not have a closed form and needs to be approximated.
Some common approximations include Laplace approximation, expectation propagation, variational inference.
See \citep{10.5555/1162254} for more details.

\section{Application: Uncertainty-based abstention}
To illustrate the utility of the GP-based uncertainty metrics, we use them to determine whether or not to trust/accept the prediction made by a given model at a given input, with the anticipation that doing so improves the precision, recall, or both at the expense of reduced prediction coverage.
We first visualize the internals of the uncertainty model and illustrate how it is used to make decisions to abstain on a UCI data set with simulated feature drift.
Next, we compare the performance of our abstaining classifier to a state-of-the-art work that works only on neural networks and is more expensive to train.

\subsection{Real data with simulated feature drift}

\begin{figure*}[h]
     \centering
     \begin{subfigure}[t]{0.45\textwidth}
         \centering
         \includegraphics[width=\textwidth]{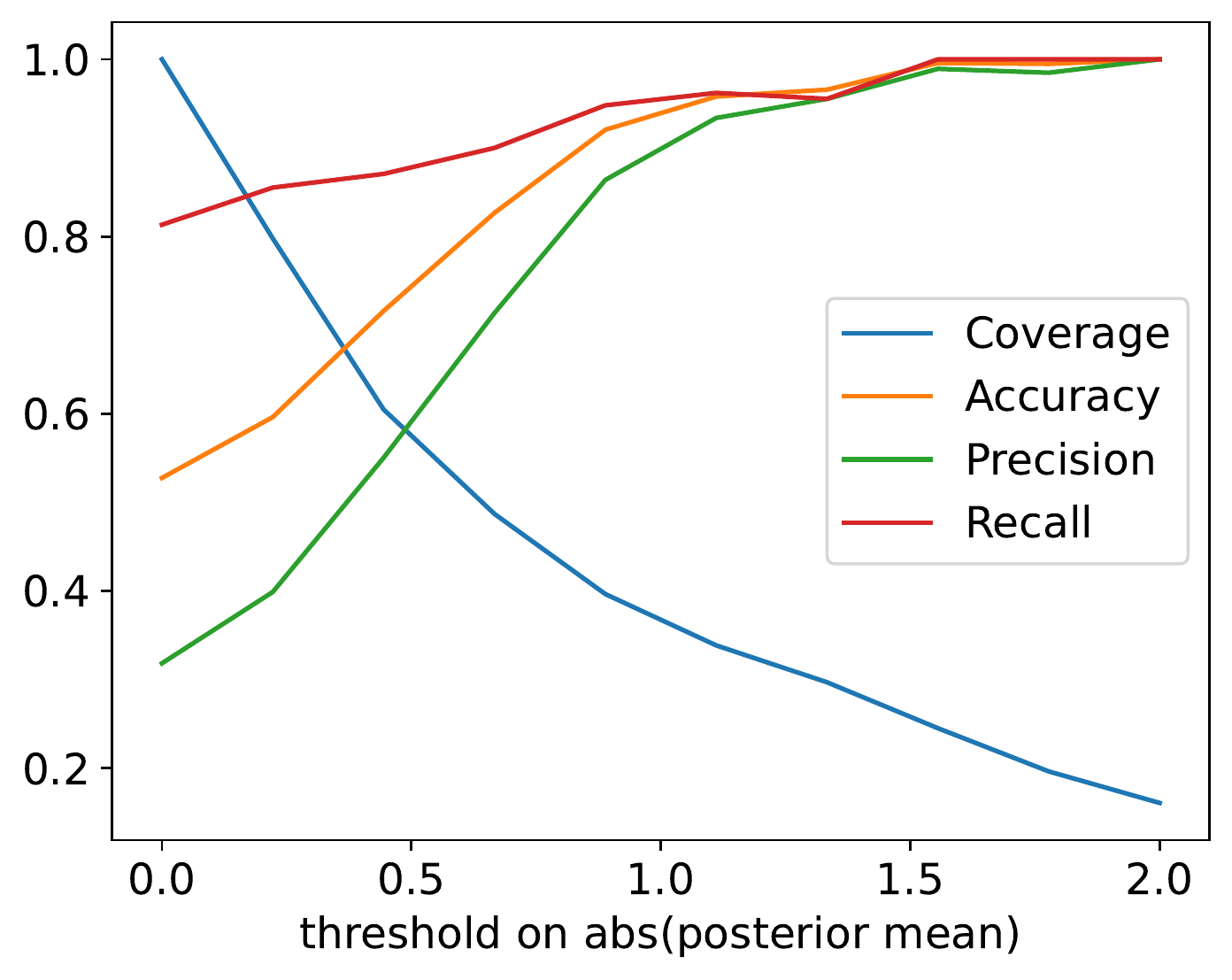}
         \caption{Abstaining when $|E[g(x)|\mathcal{D}]|<$ threshold for the reduced model using only $C1, C2$ features.}
         \label{fig:perf_mean_reduced}
     \end{subfigure}
     \hfill
     \begin{subfigure}[t]{0.45\textwidth}
         \centering
         \includegraphics[width=\textwidth]{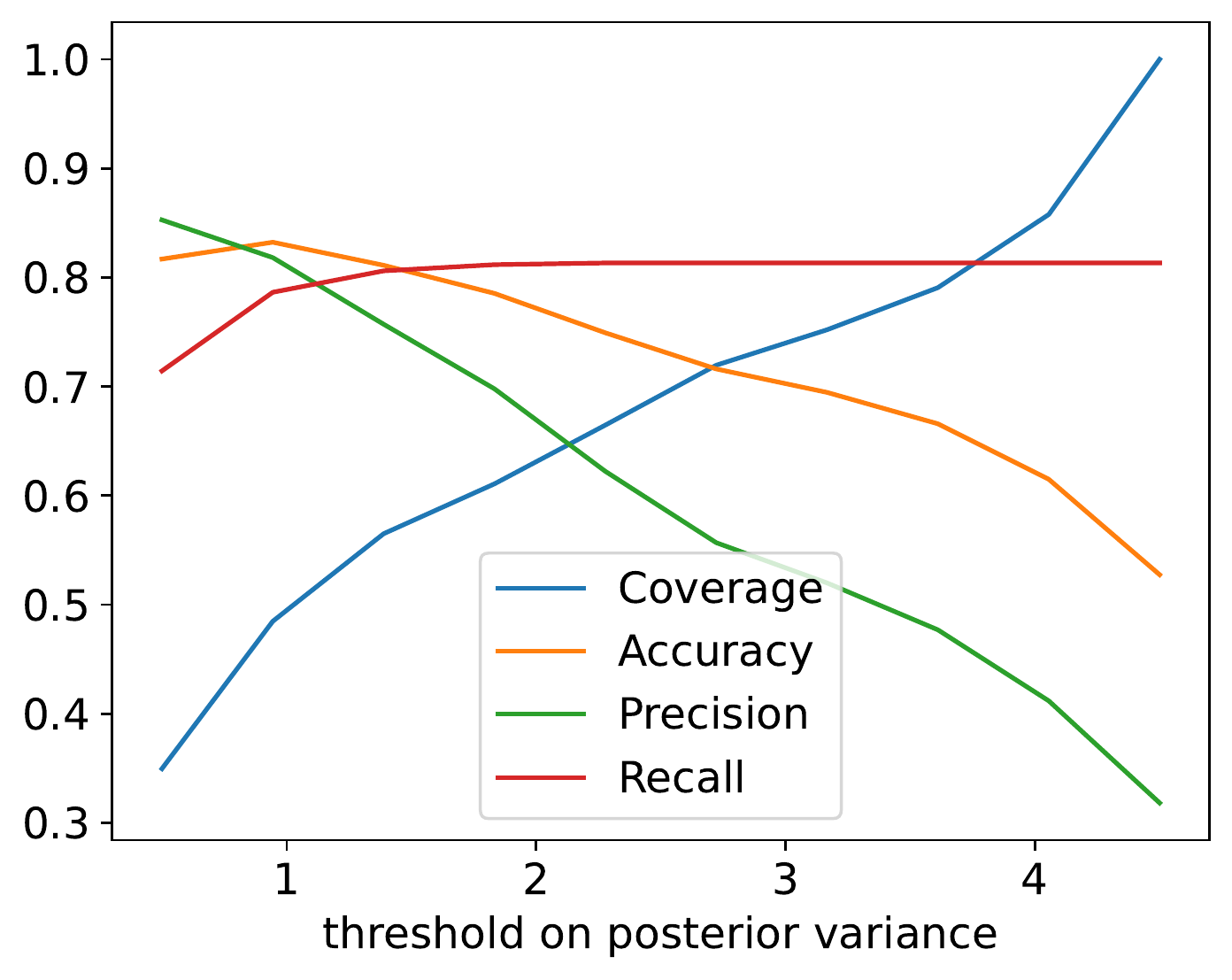}
         \caption{Abstaining when $\mathrm{Var}(g(x)|\mathcal{D})>$ threshold for the reduced model using only $C1, C2$ features.}
         \label{fig:perf_var_reduced}
     \end{subfigure}
        \caption{The tradeoff between predictive performance and the coverage as a function of the aggressiveness of abstention.}
        \label{fig:performance}
\end{figure*}
We use the Wireless Indoor Localization data set from the UCI ML repository \citep{Dua:2019} as the basis of our experiments with simulated feature drift and reduced-dimension illustrations.
There are 7 numerical features and an associated class taking one of 4 unique values.
We change this to a binary classification problem by modifying the class values to 1 for ``class is 4'' and 0 for ``class is not 4''. 
We simulate an OOD scenario by splitting the data set into different, possibly overlapping regions.
We then train the base model, train the error model, and evaluate the abstaining classifier on different regions.
The details of how the sub-data sets are constructed are as follows (See Figure \ref{fig:scatter}).
\begin{enumerate}
    \item Define two (overlapping) regions in the data: $C1 > -50$ (called ``blue'') and $C2 < -60$ (called ``yellow''), where $C1, C2$ are two of the seven features of the data set.
    \item Take a 50\% random sample from the blue region and call it $S1$. 
    \item Take a 50\% random sample from the yellow region and call it $S2$.
    \item Take a 50\% random sample from the entire data set and call it $S3$.
\end{enumerate}
The selection of features $C1, C2$ as well as the target class 4 was selected based on visual inspection of the scatter plot ensuring that the class distribution is asymmetric.
An asymmetric distribution makes it easier to simulate pronounced OOD scenarios based on sub-regions, better illustrating how our approach works.
We only use two of the seven available features so that it is easier to visualize the inner workings.

With the data sets $S1, S2$, and $S3$ defined, $S1$ is first used to train the RIPPER rule induction algorithm \citep{cohen1995fast}.
The trained RIPPER model is the following rule-based model:
\begin{align}
    \text{IF}&\nonumber\\ 
    &[C2 \leq -53.0] \lor ([C2 \leq -52.0] \land [C1 \leq -49.0])\nonumber\\
\text{THEN}&\nonumber\\
&\text{class is 4 (label = 1)}\nonumber\\
\text{ELSE}&\nonumber\\
&\text{class is not 4 (label = 0)}
\end{align}
Given that $S1$ is in the blue region, it is not surprising that RIPPER learns a decision boundary that does not depend strongly on $C1$.
With this trained RIPPER model, we compare the predictions it makes on $S2$ sampled from the yellow region against the ground truth labels and observe where the errors are made (Figure \ref{fig:error}).
A random sample of size 500 is taken from this error data to train the Gaussian process error model.
Computing the posterior distribution at a given point conditioned on the observations, 500 in our case, involves a covariance matrix of dimension 501-by-501.
This does not scale to large data sets and is one of the limitations of our method.
But we hope that we can still learn a good error model from a moderately sized data set.
For each point in $S3$, we make a prediction using the trained base model. 
Then with the trained error model, we make a decision to either accept the prediction, flip the prediction, or abstain.

The Gaussian process prior uses a squared exponential kernel with amplitude parameter set to 3 and length scale parameter set to 0.1 based on visual inspection of the resulting heat maps of Figures \ref{fig:posterior_mean} and \ref{fig:posterior_var} verifying that they look informative (e.g., not flat everywhere) but the choice was otherwise arbitrary.
The yellow points in Figure \ref{fig:abstain} where we decide to abstain from making a prediction is based on the policy of doing so if either the absolute value of the posterior mean is smaller than 1 (indicating that the entropy of the estimated Bernoulli error model is high), or if the posterior variance is larger than 3. 
If the posterior mean is a large positive value, it means that we estimate that it will usually make an error.
So we predict the opposite class in such case.

How the coverage (defined as the proportion of data points on which predictions are accepted), accuracy, precision, and recall of the model changes with respect to different abstention thresholds is shown in Figure \ref{fig:performance}.
First, we change only the threshold on the absolute posterior mean $|E[g(x)|\mathcal{D}]|$ while setting the threshold on the posterior variance large enough so that it does not contribute to any decision to abstain (Figure \ref{fig:perf_mean_reduced}).
We observe that the accuracy, precision, and recall all increase at the expense of reduced coverage, as expected.
Specifically, the accuracy increases from roughly 50\% to 80\% by abstaining on 50\% of the predictions, and increases up to 90\% by abstaining on 60\% of the predictions.
As the error model is only trained in the yellow region, we see in Figure \ref{fig:posterior_mean} that the upper region is predominantly represented by the prior mean of zero, i.e., 0.5 error rate.
Also, it learns that the trained RIPPER model fails to generalize and has a strong propensity to make a prediction error on the left part.
An abstention policy based on thresholding only the posterior variance (Figure \ref{fig:perf_var_reduced}) while setting the threshold of the posterior mean to 0 shows a similar trend.
We observe that it is only confident in the region where there is high data density (Figure \ref{fig:posterior_var}).

\begin{figure*}[h]
     \centering
     \begin{subfigure}[t]{0.45\textwidth}
         \centering
         \includegraphics[width=\textwidth]{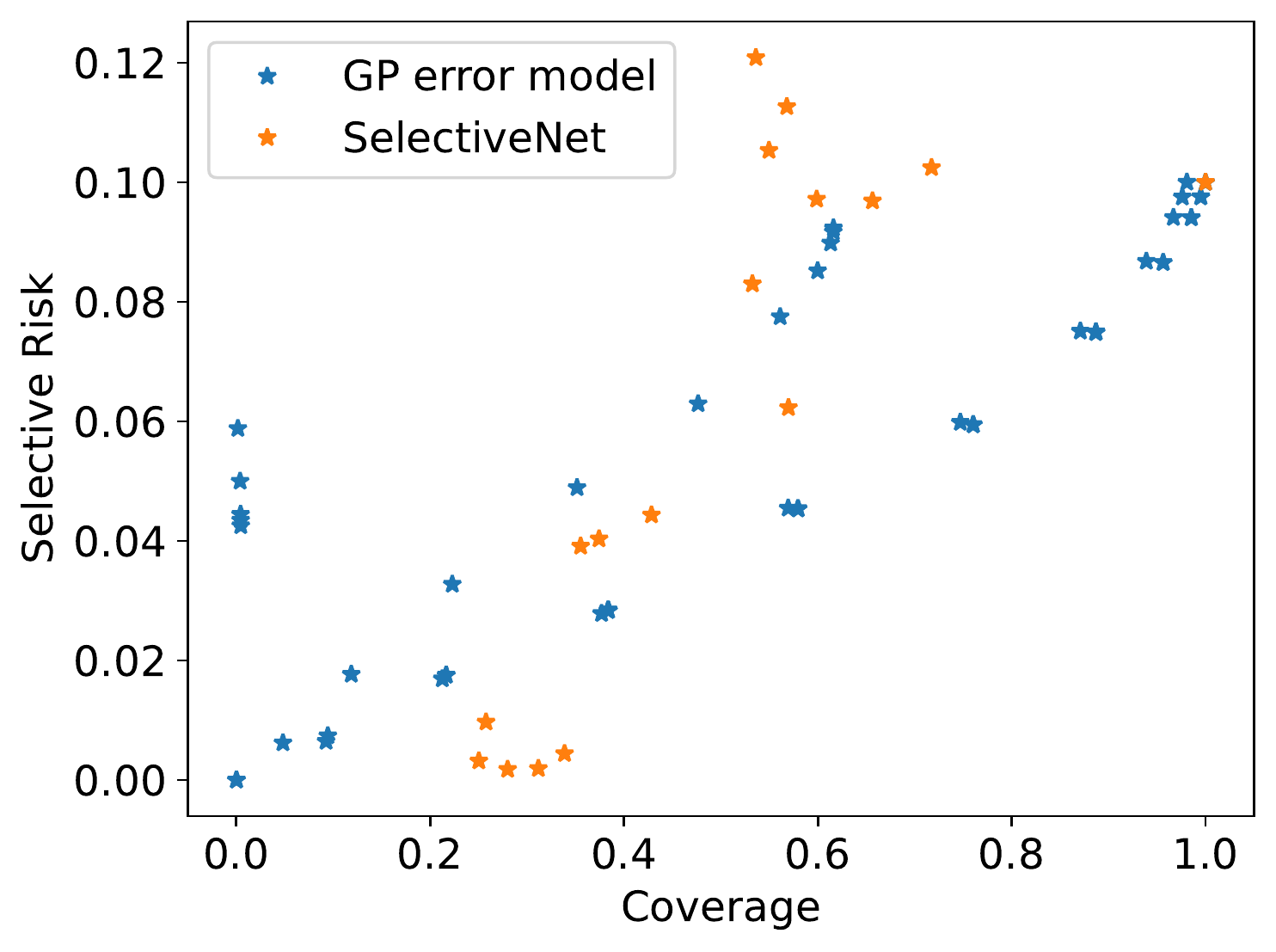}
         \caption{OOD scenario trained on dog vs. non-animals and tested on all data including animals.}
         \label{fig:gp_sn_ood}
     \end{subfigure}
     \hfill
     \begin{subfigure}[t]{0.45\textwidth}
         \centering
         \includegraphics[width=\textwidth]{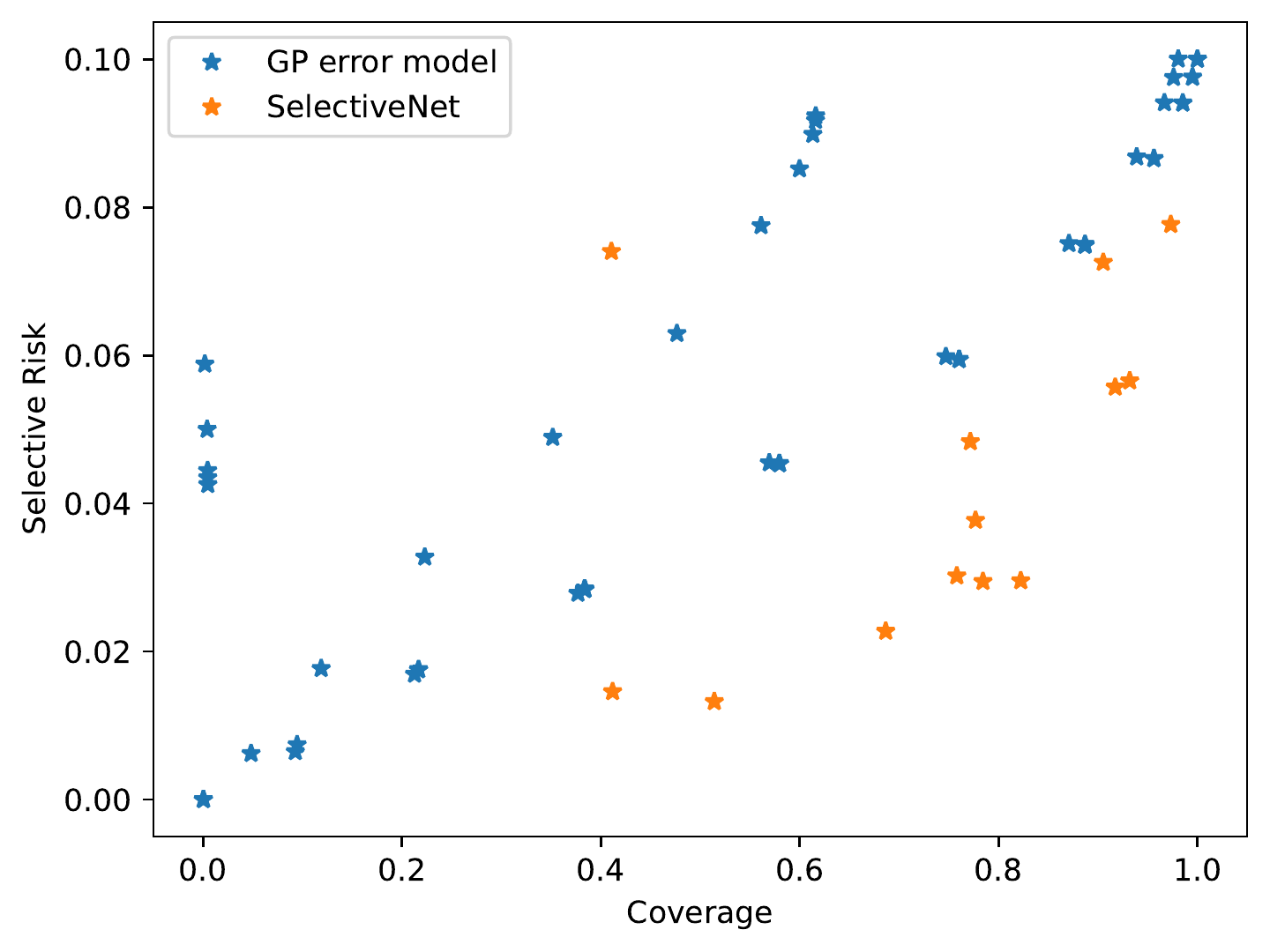}
         \caption{In-distribution scenario trained and tested on the same distribution.}
         \label{fig:gp_sn_all}
     \end{subfigure}
        \caption{The selective risk for the GP error model and SelectiveNet at different coverage levels. The GP error model was evaluated on a grid of threshold pairs on the absolute posterior mean and variance. SelectiveNet was trained and evaluated on different target coverage levels, without guarantees to achieve it.}
        \label{fig:gp_sn}
\end{figure*}
\subsection{Benchmark against SelectiveNet}

SelectiveNet \citep{DBLP:journals/corr/abs-1901-09192} is a state-of-the art abstaining classifier for neural networks.
The main idea behind it is to attach a \emph{selection head} as an output layer of the network, to inform whether to abstain, alongside the typical \emph{prediction head} which provides predictive probabilities.
The overall loss function is a weighted sum between the categorical cross entropy from the prediction head and what they call a \emph{selective risk} from the selection head.
The selective risk is defined as:
\begin{align}
    R(f, g) = \frac{E_{X,Y}[l(f(X), Y)g(X)]}{E[g(X)]}
\end{align}
where $f$ is the prediction function, $g$ is the binary-valued selection function, and $l$ is the loss function.
It is essentially the expected loss conditioned on the selected points.
The loss function we use is the 0-1 loss: $l(x, y)=\mathbf{1}(x\neq y)$.
They demonstrate that SelectiveNet outperforms MC dropout and Softmax Response in various experiments, of which includes the CIFAR-10 dataset \citep{CIFAR}.

In order to create an OOD scenario for binary classification, we use the CIFAR-10 dataset and design the experiment as follows.
The prediction target is ``dog'' vs ``not dog''.
The training set is filtered to only include those where the label is either a dog or not an animal: \{dog, airplane, automobile, ship, truck\}.
The test set is kept unmodified with all 10 labels including the animals.
First, SelectiveNet is trained with the filtered training set and the selective risk is evaluated on the test set for various coverage targets. 
Next, the base model of the GP error model, i.e., SelectiveNet without the selective head (the VGGNet), is trained on the same training set less a hold out of 1000 data points that will be used to train the GP error model.
We evaluate the selective risk of the predictions made by the base VGGNet on the OOD test set, abstaining as instructed by the trained GP error model for various thresholds on the aleatory and epistemic uncertainties.

The benchmark results are summarized in Figure \ref{fig:gp_sn}.
We observe that the GP error model is comparable to SelectiveNet for OOD scenarios (Figure \ref{fig:gp_sn_ood}), and less effective but achieves no worse than 95\% of SelectiveNet's accuracy for in-distribution scenarios (Figure \ref{fig:gp_sn_all}).
Specifically for the OOD case, the GP error model tends to out perform SelectiveNet in the high coverage regions and vice versa.
For each target coverage value, SelectiveNet needs to be retrained, so the number of experiments we could run were limited given that each data point for SelectiveNet of the plot takes 500 minutes to complete. 
On the other hand, the GP error model requires a single 500 minute training session to train the base VGGNet model, a 10 minutes session to train the GP error model, then the selected risk is reevaluated at different thresholds over $S3$.
Since this evaluation of the selective risk is cheap, we were able to create a grid of thresholds on the posterior mean and variance and evaluate the selective risk.
Considering that the GP error model is a relatively simple idea that is cheap to train (used 1000 data points) and is universally applicable, performing comparably to an abstaining classifier restricted to neural networks purposely built to optimize for selective risk, and that is expensive to train is noteworthy.

The experiments were done using a modification of the code\footnote{https://github.com/geifmany/SelectiveNet} provided by the authors of SelectiveNet.
Apart from the modification of the training set, SelectiveNet was trained with a reduced number of training epochs (10) compared to the value in the code (300), where a single epoch took roughly 50 minutes to complete on our machine.
Other parameters were left untouched.
To evaluate abstaining based on our error model, the base VGGNet was trained with almost the same data (1000 less) also for 10 epochs.
We implemented the GP regression model using the Pyro framework that approximates the posterior using variational inference.
We used a squared exponential kernel with length scale set to 20 and variance set to 3.
The choice of variance was arbitrary, and the length scale was chosen after observing the range of the posterior mean and variance values to make sure it is large enough to show some variability.

\section{Conclusion}
Quantifying the uncertainty of predictions made by rule-based models, or other inherently binary models cannot take advantage of the large body of work on calibration, conformal prediction, or purpose built architectures for neural models.
Existing meta-models applicable to rule-based models do not support providing confidence metrics for individual inputs, or are not designed to work in OOD scenarios.
We presented a solution to the problem of quantifying uncertainty of the correctness of prediction for binary classification models based on the simple idea to directly model its error rate over the entire input space.
This approach is universally applicable to any binary classification model through a plug-in interface, and it does not depend on any of the implementation details of the base model.
We demonstrated that it exhibits ``common sense'' behavior such as being uncertain at the class boundaries and in regions where observations are sparse.

The most apparent limitation to our approach is being restricted to binary classification.
Other limitations involve the choice to model the error rate as a Gaussian process.
There are well known scalability problems regarding how the size of the data affects the dimensionality of the conditional normal distribution.
Like any Bayesian model, the need to tune the hyperparameters, and the very fact that our uncertainty metrics are in the end beliefs that depend on our choice of the prior rather than objective probabilities are also notable drawbacks.
Nevertheless, we have demonstrated that such a simple, cheap, and generally applicable idea is comparable to state-of-the-art techniques that have limited scope of application and are more expensive.

Embracing the fact that AI systems are far from perfect, and trying to shed some light onto when they are unsure or wrong is an indispensable step towards broader acceptance and trust in involving AI in our society and lives.
Through providing a machine with the capability to say ``I'm not sure'' is a first step towards building systems that have a basic level of situational- and self-awareness. 
Efforts to advance such capabilities is a much needed counterbalance to approaches that exclusively focus on performance metrics. 


\section*{Acknowledgments}
This work has been partially funded by the French government as part of project PSPC AIDA 2019-PSPC-09, in the framework of the "Programme d'Investissement d'Avenir". 

\bibliographystyle{abbrvnat}  
\bibliography{biblio}

\end{document}